# A Zipf's Law-based Text Generation Approach for Addressing Imbalance in Entity Extraction

**Zhenhua Wang[1], Ming Ren[1*], Dong Gao[2], Zhuang Li[2]**

[1]School of Information Resource Management, Renmin University of China, Beijing, China
[2] College of Information Science and Technology, Beijing University of Chemical Technology, Beijing, China
zhenhua.wang@ruc.edu.cn; renm@ruc.edu.cn; gaodong@mail.buct.edu.cn; 2021210511@mail.buct.edu.cn

**Abstract:** Entity extraction is critical in the intelligent advancement across diverse domains. Nevertheless, a challenge to its effectiveness arises from the data imbalance. This paper proposes a novel approach by viewing the issue through the quantitative information, recognizing that entities exhibit certain levels of commonality while others are scarce, which can be reflected in the quantifiable distribution of words. The Zipf's Law emerges as a well-suited adoption, and to transition from words to entities, words within the documents are classified as common and rare ones. Subsequently, sentences are classified into common and rare ones, and are further processed by text generation models accordingly. Rare entities within the generated sentences are then labeled using human-designed rules, serving as a supplement to the raw dataset, thereby mitigating the imbalance problem. The study presents a case of extracting entities from technical documents, and experimental results from two datasets prove the effectiveness of the proposed method. Furthermore, the significance of Zipf's law in driving the progress of AI is discussed, broadening the reach and coverage of Informetrics. This paper presents a successful demonstration of extending Informetrics to interface with AI through Zipf's Law.

## 1. INTRODUCTION

Entity extraction aims to identify and reveal various entities mentioned in documents, such as locations and places, which allows us to deeply understand and easily explore the knowledge and insights embedded in sources like the web, digital libraries, technical documents. Esteemed for its indispensability, entity extraction has widespread applications and exerts a profound influence across multiple domains. For instance, it can be used to build industry safety knowledge graph (Wang et al., 2022), uncover publicly unknown knowledge (Song et al., 2015a), enhance patent similarity measurement (An et al., 2021), retrieve knowledge on waste reuse from patent reports (Spreafico and Spreafico, 2021), examine the relationship between drug and side effect (Jeong et al., 2020), and identify algorithms (Wang and Zhang, 2020), etc. (Song et al., 2015b).

Great efforts have been dedicated to entity extraction, and currently deep learning-based methods have emerged as the dominant choice due to their remarkable advantages in scalability and accuracy. However, in certain domains, entity extraction faces a challenge related to data imbalance. To be specific, some categories of entities are common, while others are scarce and dispersed, which has a significant impact on the model's performance. Unfortunately, manually labeling additional data on a small scale to address this issue is both time-consuming and labor-intensive. It is also hard to handle through loss function assignment and re-sampling, since they weakly assign additional excitation (Johnson and Khoshgoftaar, 2019).

An approach worth exploring is text generation, which has gained widespread attention for its ability to autonomously produce text that closely resembles human writing (Anaby et al., 2020). One prominent language model is GPT2, which has found extensive applications in various generative tasks like text-to-speech synthesis (Saeki et al., 2021), event detection (Veyseh et al., 2021), brainwave opera (Pearlman, 2021), etc. GPT2 employs a training strategy based on predicting the next word and utilizes conditional probability modeling to generate subsequent sequences based on known input sequences (Anaby et al., 2020). This technique is beneficial for generating text pertaining to a greater number of small-scale entities at an acceptable cost, which can then be extracted using designed rules, thereby mitigating the volume disparity among entities. However, a limitation is its tendency to generate text with a random orientation, which inadvertently favoring frequently-used entities rather than providing adequate emphasis on the less prevalent small-scale entities.

Such apprehension can redirect inquiries towards Informetrics, where the theory of text is a widely recognized domain. Within this context, a hope emerges for addressing this engineering issue – the prospect of employing quantitative analysis, particularly in relation to words. At the representative of this notion stands Zipf's Law, a statistical phenomenon that describes the frequency of a word as inversely proportional to its rank (Newman, 2005; Piantadosi, 2014). It provides valuable insights into the distribution of word frequencies, and can serve as a guiding principle for developing effective strategies to address data imbalance in entity extraction tasks. These strategies may involve the development of specialized processing techniques for handling texts that encompass common entities and those that encompass fewer, respectively. To be specific, words can be divided into commonly used ones and rare ones based on their frequency. Generally speaking, texts related to rare entities tend to include more rare words. Therefore, we select sentences containing more rare words (i.e. rare sentences) for text generation, in order to obtain more texts related to rare entities. In addition, some rare entities may also be composed of common words or used in conjunction with common words, so we also use sentences containing more common words (common sentences) to generate refined text to integrate relevant information without diluting rare entities.

This paper proposes a **Zi**pf's law-based **te**xt **g**eneration approach for entity extraction (Ziteg). Given documents from a specific domain, Ziteg first utilizes Zipf's law to classify the words into two categories: common and rare. Then a sentence discriminant is developed to further categorize the sentences as common or rare ones. To handle these sentences separately, we train two distinct GPT2 models. Rare entities present in the generated sentences are automatically labeled using predefined rules, and are added to the dataset as supplementary data. Thus, the texts related to various categories of entities are relatively balanced, effectively improving the performance of entity extraction. Our novel reconsideration of Zipf's law amplifies Informetrics from theory to practice. Intriguingly, this exploration also serves as a preliminary investigation into the Zipfian behavior within entities themselves. This not only magnifies the influence of Informetrics but also expands its intellectual frontiers, further enriching its realm of knowledge. We hold an affirmation that Informetrics assumes a positive drive in shaping the trajectory and progress of AI.

The main contributions of this study are threefold.

(1) Zipf's law is exploited for text generation to address the imbalance challenge in entity extraction.

(2) A novel approach Ziteg is proposed and an application is presented to extract entities from technical reports. The effectiveness of Ziteg is demonstrated through extensive experiments.

(3) The discussion encompassed both the Zipf's Law itself and its implications for AI research. It highlighted the potential of applying Informetrics theory in practical scenarios and leveraging its synergy with AI, confirming that Informetrics can emerge as a prominent player in current era.

The rest of the paper is organized as follows. Section 2 introduces the related work on entity extraction, data imbalance and Zipf's law. Section 3 and 4 present Ziteg and an application case in detail. A series of experiments and results are presented in Section 5 with discussions in Section 6. The paper is concluded in Section 7.

## 2. RELATED WORK

### 2.1. Entity extraction

Entity extraction, also known as named entity recognition, lays the foundation for various natural language processing tasks, such as knowledge graph, question-answering and machine translation (Liu et al., 2022; Wang et al., 2022b), as well as a variety of fields, such as healthcare, chemistry, biomedicine, cyber security, news media and finance (Song et al., 2015a; Song et al., 2015b; Jeong et al., 2020; Wang and Zhang, 2020; An et al., 2021; Spreafico and Spreafico, 2021; Chen et al., 2022; Gomes et al., 2022; Wang et al., 2022a; Zheng et al., 2022).

Given a piece of document as a sequence of tokens $S = <c_1, c_2, …, c_n>$, a list of entities is identified in the form of $<I_b, I_e, k, t>$, where $k$ refers to an entity, $I_b$ and $I_e$ enumerates the entity span from $I_b$ to $I_e$, $t$ refers to the entity type (Liu et al., 2022; Wang et al., 2022b). For example, from the content "*The low temperature of fuel gas may cause backfire of tail gas incinerator*", we have the output as {$c_4, c_5$, fuel gas, Material; $c_{10}, c_{12}$, tail gas incinerator, Equipment; $c_1, c_2$, low temperature, State; $c_8, c_8$, backfire, Consequence}, which contains 4 categories of entities.

Earlier works of entity extraction mainly focused on grammar rules and dictionaries, which are relatively simple but hard to adapt to other projects. Then the studies gradually shifted to traditional machine learning-based methods, which accomplish entity extraction as a sequence labeling task using SVM and HMM, etc. The coding schemes B-I-O is widely used that assign tags to tokens indicating their role in a named entity, where B represents the beginning, I represents inside and O represents outside. Nowadays, the deep learning-based methods have emerged as the dominant paradigm due to the ability to automatically learn and extract features to make accurate predictions (Liu et al., 2022). They typically employ an architecture consisting of an embedding layer (e.g., BERT), an encoding layer (e.g., CNN and BiLSTM) and a decoding layer (e.g., CRF). This framework has yielded significant advancements across various tasks. For instance, BERT-CRF has demonstrated promising results in Portuguese named entity recognition (Souza et al., 2019), while LSTM and CRF have proven effective in identifying biomedical named entities (Habibi et al., 2017). Additionally, BiLSTM-CRF has facilitated the extraction of entities such as place and person names (Huang et al., 2015). Nonetheless, deep learning-based approaches demand substantial effort in labeling training data, and the model's performance typically improves with larger labeled datasets. Nevertheless, they often struggle to address the issue of imbalanced data in specific domains, thereby affecting the overall performance.

## 2.2. Data imbalance

The data imbalance presents a common challenge in classification tasks and can significantly impact the model's performance (Johnson and Khoshgoftaar, 2019). The approaches to address data imbalance can generally be categorized into two tracks. The first track focuses on modeling the loss function, where various techniques have been employed to assign weights to different classes, e.g., class-weighted loss and focal loss commonly used in computer vision (Lin et al., 2017; Deepak, S., & Ameer, 2023).

The second track centers around on the data itself, which has gained popularity due to its greater controllability and practicality (Anil and Singh, 2020; Chowdhury, 2021; Martinez et al., 2021). The common methods are under-sampling and over-sampling (Mohammed et al., 2020). Over-sampling involves randomly selecting rare-class samples with replacement multiple times during each training batch to balance the sample sizes. Under-sampling involves randomly removing some majority-class samples during training.

There is also a challenge of entity imbalance in tasks, such as extracting chemical and biomedical entities (Akkasi et al., 2018), social media entities (Peng and Dredze, 2016), and entities from electronic patient records (Grancharova et al., 2020). Entity imbalance refers to the uneven distribution of entities within a specific domain or entity type. Addressing these challenges solely through sampling methods is ineffective. Over-sampling may lead to model overfitting due to repeated samples, while under-sampling may remove crucial information that could influence model training. Furthermore, traditional sampling methods fail to introduce new stimuli or provide additional diversity beyond the original dataset, as the messages within documents are complex and interconnected, and the content varies across different documents. Consequently, the potential for improvement is limited.

## 2.3. Zipf's law

Zipf's law initially describes the phenomenon where the frequency of a word in a natural language corpus is inversely proportional to its rank, denoted as $F \cdot R = C$ ($C$ is a constant), thereby establishing a deterministic link between them (Newman, 2005; Piantadosi, 2014). Generally, there are two frequent forms of Zipf's law on natural language, i.e., the classical Zipf's law and extended one, which formulate the interaction between word frequency and its rank in different manners. The curve fitted by Zipf's law can be dichotomized into the head part and tail part, with the head part more related to common words and the tail part to rarely used words (Wang, 2021; Xiao and Zeng, 2022).

What is essential in all those observed Zipf laws or Zipf-like models is the decreasing tendency of rank distributions, more or less close to a straight line in log-log plot, which must have some underlying universal mechanisms, being generic and common to all (living) systems considered (Wang, 2021). From the perspective of individual words, the observed distribution can be characterized as an imbalance shaped by the coexistence of high-frequency words and low-frequency ones. This asymmetry can be attributed to the limited human cognitive resources, i.e., our cognitive capacity imposes restrictions on the frequency and usage of words, resulting in an uneven distribution.

Zipf's law has been found to apply to many other types of data studied in a wide array of disciplines, including user-chosen passwords (Xiao and Zeng, 2022), speech acts (Qi and Wang, 2023), classical music (Serra et al., 2022), etc. Zipf's law has also extended influence in diverse domains such as human interactions, natural disaster prediction, and pollution emission control (Rossi et al., 2013; Wei et al., 2021; Valero et al., 2022).

## 3. **METHOD**

The architecture of Ziteg consists of three modules, the word classification module, the sentence classification and generation module, and the rare entity labeling module, as depicted in Fig.1.

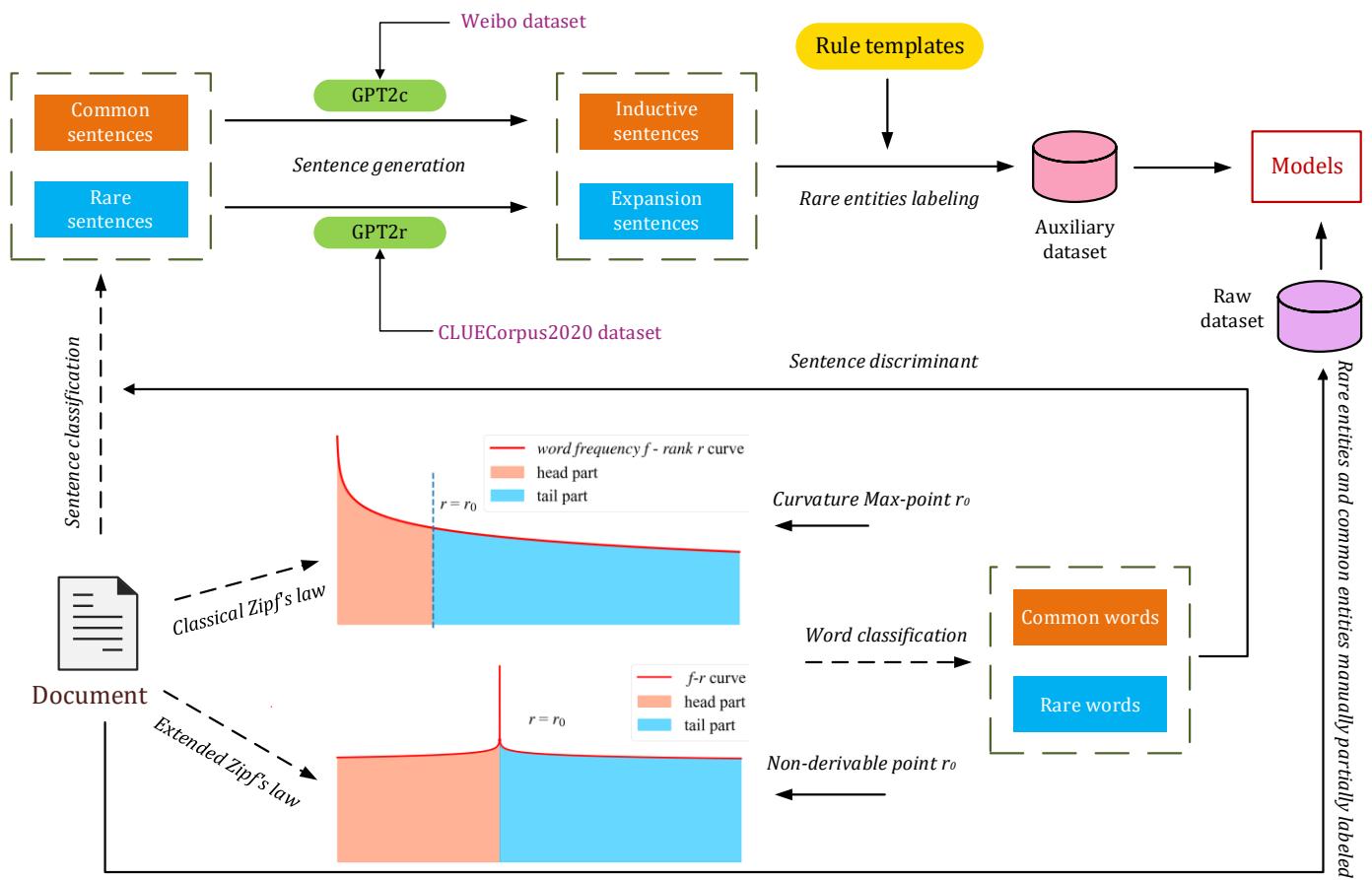

Fig.1: The architecture of Ziteg.

### 3.1. Word classification by Zipf's law

The words in the documents are classified into common and rare ones using Zipf's law. The two forms of Zipf's law, i.e., the classical Zipf's law and extended one, are both used in this study.

#### 3.1.1. Modeling by the classical Zipf's law

The classical Zipf's law models the *f-r* as Eq.1,

$$f(r) = \frac{C}{r^{\alpha}} \tag{1}$$

where $\alpha$ and $C$ are constants determined by the type of document, and can be estimated using linear regression. Fig.2 shows the classical Zipf distribution curve, and the point $r_0$ that distinguishes the head part and tail part needs to be found in order to classify words. Since there is no obvious turning point or tangent one in this curve, the most inflection point in the curve is taken as $r_0$.

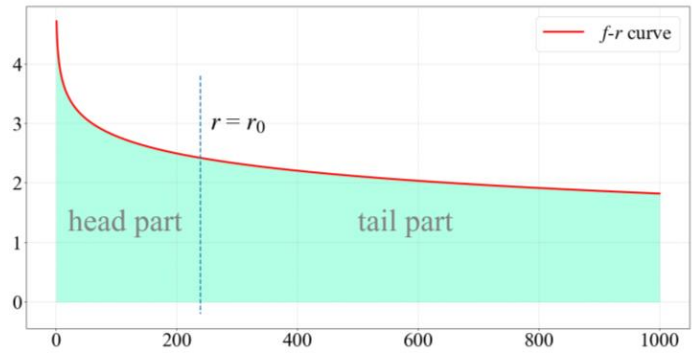

Fig.2: The classical Zipf distribution.

Fortunately, the curvature that reflects the bending degree of the curve meets our requirements, and the maximum point of the curvature function is taken as $r_0$. Fig.3 shows an example of the curvature curve.

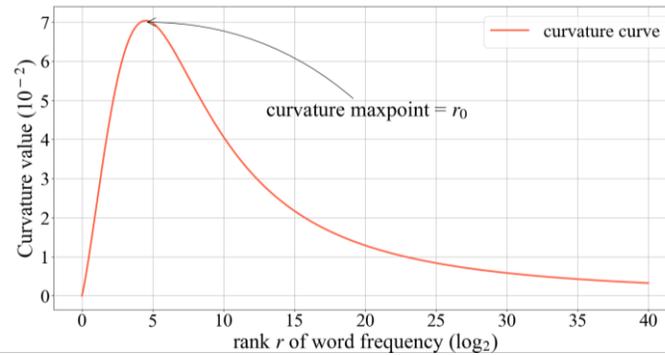

Fig.3: An example of the curvature curve (log-scaled).

The curvature function $\kappa(x)$ of the curve $y$ is Eq.2.

$$\kappa(x) = \frac{\mid y'' \mid^{3/2}}{[1 + (y')^2]} \tag{2}$$

For the classical Zipf distribution, the curvature function $\kappa(r)$ can be obtained through Eq.3.

$$\kappa(r) = \frac{C\alpha(\alpha+1)r^{-\alpha-2}}{[1 + (C\alpha \cdot r^{-\alpha-1})^2]^{\frac{3}{2}}} \tag{3}$$

And $r_0$ can be calculated as Eq.4.

$$r_0 = \arg\max \ \kappa(r) \tag{4}$$

Obviously, $r_0$ is the only effective curvature extreme point by analyzing the derivative of Eq.3 ($C > 0$ and $\alpha > 0$). Based on $r_0$ ($r_0 > 0$), the words with the rank from 1 to $r_0$ are classified as common ones, and the rest as rare ones. They refer to the words used relatively more frequently and those used less frequently.

### 3.1.2. Modeling by the extended Zipf's law

The extended Zipf's law can be represented by Eq.5,

$$f(r) = \frac{C}{(r+\beta)^\alpha} \tag{5}$$

where $\alpha$, $\beta$ and $C$ are constants that vary depending on the document type and can be estimated using linear regression. Fig. 4 illustrates the extended Zipf distribution curve, which exhibits a turning point at $r_0 = -\beta$. This turning point serves as a threshold to differentiate between the head part and tail part. Words with ranks between 1 and $r_0$ are categorized as common ones, while those beyond $r_0$ are considered rare ones.

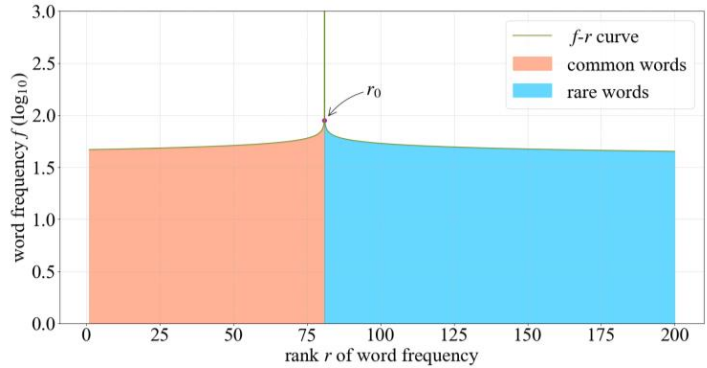

Fig.4: Illustration on extended Zipf's law.

### 3.2. Sentence classification and generation

This module classifies sentences as common and rare ones, and then build two GPT2 models for text generation.

### 3.2.1. Sentence classification

For each sentence $s$ containing $w_h$ words, compute the $\tau$ value using Eq.6, which compares the proportion of common words $w_c$ to all words to the proportion of $r_0$ to total rank $r_t$. If the $\tau$ value is above 0, the sentence is considered more relevant to common words and classified as a common sentence ("1"). Otherwise it is classified as a rare sentence ("0"). This approach adheres to the Occam razor principle, emphasizing conciseness and consistency.

Algorithm 1 outlines the overall process of sentence classification.

$$s = \begin{cases} 1, & \tau > 0 \\ 0, & \tau \le 0 \end{cases}; \ where, \ \tau = \frac{w_c}{w_h} - \frac{r_0}{r_t} \tag{6}$$

| Algorithm 1: Sentence classification via Zipf's law |
|---|
| **Input**: All sentences $S$ in documents |
| **Parameter**: All words $W$ in descending order of frequency; Eq.3-6. |
| **Output**: common sentence $S_c$, rare sentence $S_r$ |
|   1:  $w_c = \varnothing,\ S_c = \varnothing,\ S_r = \varnothing$ |
|   2.  **Case** I: // classical Zipf's law |
|   3:    $\kappa(r) \leftarrow$ Eq.3 |
|   4:    $r_0 \leftarrow$ Eq.4 |
|   5:  **Case** II: // extended Zipf's law |
|   6:    $r_0 \leftarrow$ Eq.5 |
|   7:  **for** $i \leftarrow 1$ to $r_0$ **do** |
|   8:     $w_c = w_c \cup W[i]$ |
|   9:  **for** each sentence $s$ in $S$ **do** |
| 10:     $s \leftarrow$ Eq.6 |
| 11:     **if** $s = 1$ **do** |
| 12:       $S_c = S_c \cup \{s\}$ |
| 13:  $S_r = S - S_c$ |
| 14:  **return** $S_c, S_r$ |

### 3.2.2. Sentence generation

For text generation, we employ GPT2 and build two models: GPT2r and GPT2c. GPT2r produces expanded sentences based on rare sentences, and GPT2c induces sentences based on common sentences. Note that GPT2 is a tool, and evaluating its performance is beyond the scope of this study.

#### 3.2.2.1 Rare sentence generation

In cases where the training data for GPT2r is insufficient in certain domains, an open dataset CLUECorpus2020 that contains approximately 5 billion words can be used jointly. In data processing, each article in the dataset is marked with "MASK" at the beginning, "CLS" is used to distinguish different articles, and "SEP" is inserted between lines. Please refer to Eq.7, where $a$ represents the article, and $l$ represents the paragraph line.

$$u = | MASK | l^1_1 | SEP | l^1_2 \cdots | SEP | l^1_n | CLS | \cdots | MASK | l^a_1 | SEP | l^a_2 \cdots | SEP | l^a_m | CLS | \qquad (7)$$

For the corpus, GPT2r constructs a series of tokens connected by token embedding mapped by BERT's Tokenizer and its position embeddings. and maximizes the likelihood function of Eq.8 through the *Transformer* decoder with 12 parallel layers and *attention* with 12 parallel heads.

$$Loss = -\sum \log P(\omega_i \mid \omega_{i-k}, \ldots, \omega_{i-1}; \theta) \qquad (8)$$

where $k$ represents the size of the sliding window that predicts the current token $w_i$ conditioned on $k$ historical tokens $\{w_{i-k}, \ldots, w_{i-1}\}$, $\theta$ refers to the parameters of the network, and uses Adam to optimize the likelihood function. The learning rate of 0.001, the batch with size of 8, GELU activation function, etc., are used.

GPT2r implements unsupervised training, and employs an auto-regressive $n$-gram method to predict words sequentially. GPT2r trained by 10 epochs is called to perform sentence expanded prediction, and for each rare sentence two expanded sentences with a length of 500 are generated.

### 3.2.2.2 Common sentence generation

Many specific domain sentences often consist of technical terms and exhibit a consistent logical structure, but they may lack linguistic diversity. Besides, there is no corresponding labels. Fortunately, *Weibo* contains discussions on diverse topics, which often carry titles when recording, commenting, and forwarding messages. These titles may serve as suitable labels for training GPT2c. Although the inductive sentences may be vague or irrelevant, their presence does not impact the labeling of rare entities.

To do so, raw data is crawled from *Weibo* and undergoes a cleaning process that involves removing duplicates, emojis, "##" symbols, and other unnecessary elements. Data items with less than 100 words and titles containing less than 2 words are discarded, since they lack sufficient information. Finally, Eq.9 is employed to construct the training dataset, which consists of 24,000 instances.

$$\{(x_i, y_i)\}_{i=1}^n = y_1 | SEP | x_1 | EOS | y_2 | \cdots | y_n | SEP | x_n | EOS \qquad (9)$$

where $x_i$ represents the title, $y_i$ represents its content, *SEP* is used to separates the title and content, and *EOS* is used to mark the end of a sample, allowing the model to accept the next one.

We apply BERT's Tokenizer to embed the dataset to form token embeddings, and shape position embeddings based on the position of each token. To better distinguish between the content and title, *SEP* is embedded as segment embeddings. The concatenation of the three is fed to the *Transformer* decoder with 6 parallel layers and *attention* with 12 parallel heads, and the conditional probability of the known sequence is modeled by autoregression to predict the subsequent sequence. The cross-entropy loss function is undertaken to only minimize the loss value of the title part for training. The optimizer is Adam with learning rate of 0.0001, batch size is 16, activation function is GELU. GPT2c with 10 epochs of training is called to generate one inductive sentence for each common sentence.

### 3.3. Rare entity labeling via rules

The generated sentences, including the expanded sentences and inductive ones, encompass additional rare entities, which are usually less intricate compared to the sentences found in the original documents. Thus, it is suggested to label these rare entities via rules with certain feasibility (Vlachidis and Tudhope, 2016; Yang et al., 2021) and use them as supplement to mitigate the imbalance problem. Note that the focus is on ensuring the accuracy of the labels rather than the quantity, in order to minimize the introduction of additional errors. Different domains often require different sets of rules. In this paper, relevant guidelines are provided to assist in this regard.

***Part-of-speech (POS)-dominated rule template***: POS labels are determined by the expression and recording logic used in the sentences. When dealing with domain-specific entities that have descriptive characteristics, specific POS collocations can be formulated for labeling purposes. For instance, in domains like archaeology, a rule can be:

$$\{NNP \cup NN \cup VBN \cup IN \cup NNP\},$$

where *NNP*, *NN*, *VBN* and *IN* stand for proper noun, normal noun, past participle and preposition / subordinating conjunction, which helps in labeling entities in e.g., "coin associated with hearth spot."

***Regular expression-dominated rule template***: This rule template uses descriptive formulas to define entities based on specific matching patterns of strings, and is particularly suitable for the domain-specific entities that involve numbers, symbols, emojis and similar elements. For example, consider the following designed rule template:

$$[A-z]or + [A-Z]([1-9]\backslash d* \mid 0)(\backslash. \backslash d+) ? , ([1-9]d* \mid 0)(\backslash. \backslash d+)or[A-z]([A-Za-z]+, [A-Za-z]+),$$

it can be used to label elementary mathematics entity such as "Y, Y(1, 2), Y(a, b)".

***Keyword matching-dominated rule template***: This rule template focuses on labeling entities that do not have a specific form by performing strong matches against specific words. For example, in certain domains where entities indicating intention needs to be labeled, a keyword matching sequence can be designed as follows:

$$\wedge [Who] \$books\$[*]\$ [What] \$from\$ [Where] \$to\$ [Where] \$[*]\$ [Time] *\$,$$

which enables the identification of customer needs similar to the sentence "Johnson books a flight from Washington to London tomorrow." In this template, the "^" symbol denotes the prefix identifier, indicating the beginning of the string. The "*" symbol represents any character that can match any number of characters. The "$" symbol signifies the identifier at the end of a word or phrase. The words "books" "from" and "to" are considered equivalent to mathematical constants, while "Who" "What" "Where" and "Time" correspond to mathematical variables.

***Word dependence & grammar-dominated rule template***: This rule template incorporates the relationships between words based on POS, including subordination, coordination, verb-object, and subject-predicate relationships. Additionally, it considers word length as a supplementary criterion. This approach is particularly applicable in industries such as process, civil engineering, construction and other industries. For instance, a rule template can be designed as follows:

{subordination; verb-noun; subject-verb; word length < 8}

which is utilized to label industry safety entities such as "oil-spill incident". It takes into account the relationships between words such as subordination, verb-noun, and subject-verb, along with the condition that the word length should be less than 8.

***Semantic web rule language-dominated rule template***: This rule template is based on ontology and semantics and has extensibility. For example, in the domain of medicine, a rule template can be designed as follows:

The body (antecedent): Patient (?x) ^ make (?x, ?y) ^ COVID-19 (?y, ''side"); The head (consequent): Pass in (?x, isolated).,

which can effectively match entities in sentences like "patients infected with COVID-19 must be isolated".

***Other***: It is particularly useful in domains that possess a clear sentence layout. In the context of labeling reference meta entities, a rule template can be employed to capture references like "Ferrer-i-Cancho, R., & Vitevitch, M. S. (2018). The origins of Zipf's meaning frequency law. JASIST, 69(11), 1369-1379.", as follows.

$[Aw]_2 : [Yw]_1 : [Tw]_7 : [Jw]_1 : [V] : [I] : [Pg]$,

where $Aw$, $Tw$, $Jw$, $V$, $I$, $Pg$ are the abbreviations of author words, title words, periodical words, volume, issue, and pagination respectively, and subscript indicates the quantity of each item.

It is worth noting that auxiliary trigger words can be used to verify the results. The characters in the sentences are labeled using the BIO scheme.

## 4. CASE

This section presents a case of extracting entities from technical documents termed Hazard and Operability Analysis (HAZOP) reports. The HAZOP technique is developed to identify risks and evaluate safety in various industry systems (Standard and IEC61882, 2021). It has become an increasingly indispensable paradigm, particularly with the growing volume of national development. In China, a great quantity of official policies and government statements have been issued, mandating the implementation of HAZOP for every industry process before production and development commence. HAZOP takes the nodes of the system under study (e.g., electrolytic cell, stripper) as the starting point and analyzes hazards from the perspectives of causes, deviations, consequences, suggestions and mitigation measures. The results are documented in HAZOP reports, which contain a wealth of industry safety entities, enjoying practical and promising applications (Wang et al., 2021; Wang et al., 2022a; Wang et al., 2023; Zhang et al., 2023). However, existing studies have overlooked the issue of data imbalance, which to some extent affects the model's performance. Since HAZOP reports are confidential and accessible only to personnel involved in the HAZOP project, the available HAZOP data is usually limited, making text generation a beneficial approach.

Regarding the entity categories found in HAZOP reports, there are primarily four categories: equipment (e.g., Fischer Tropsch reactor), material (e.g., hexamethylene diisocyanate), consequence (e.g., excessive liquid level), and state (e.g., insufficient oxygen content). The imbalance problem can be observed in two aspects. Firstly, a material or equipment often exhibits multiple states and consequences within a process. For example, a Fischer Tropsch reactor may encounter surge, abnormal noise and carbon deposition, etc. Secondly, while the types of equipment and materials may differ across processes, the states and consequences often remain consistent. For instance, the materials and equipment used in natural gas exploitation and rubber production processes may vary significantly, but they may encounter similar or even identical states such as low liquid level and high temperature. Consequently, entities in the equipment and material categories tend to be rare and scattered, whereas others are more common. Overall, this scenario exemplifies a typical case of entity imbalance.

### 4.1. Document description

As arranged by Brunei group, Sichuan Petrochemical, Liaoyang Petrochemical, Shenhua Group and Yanshan Group, HAZOP assessments was launched for 13 processes that are of great significance for environmental protection and energy sustainability.

Table 1: Information about HAZOP reports

| Type | Process | Volume of words | $r_t$ |
|------|---------|-----------------|-------|
| Petroleum | diesel hydrogenation<br>solvent regeneration<br>heavy oil catalytic cracking<br>naphtha isomerization | 278910 | 5416 |
| sulfur | sulfur recovery<br>desulfurization and sulfur foam<br>sulfur production | 306147 | 6024 |
| coal | indirect coal liquefaction<br>coal washing<br>coal cracking | 135460 | 5070 |
| gas | formic acid to carbon monoxide<br>water electrolysis<br>ammonium nitrate to nitrous oxide | 242173 | 6524 |

The assessments resulted in 13 HAZOP reports that are categorized into four types: petroleum, sulfur, coal and gas, see Fig.1. Fig.5 displays the frequency-rank curves for each report type. These curves illustrate a closely linear relationship between word frequency and rank on a logarithmic scale, consistent with Zipf's law.

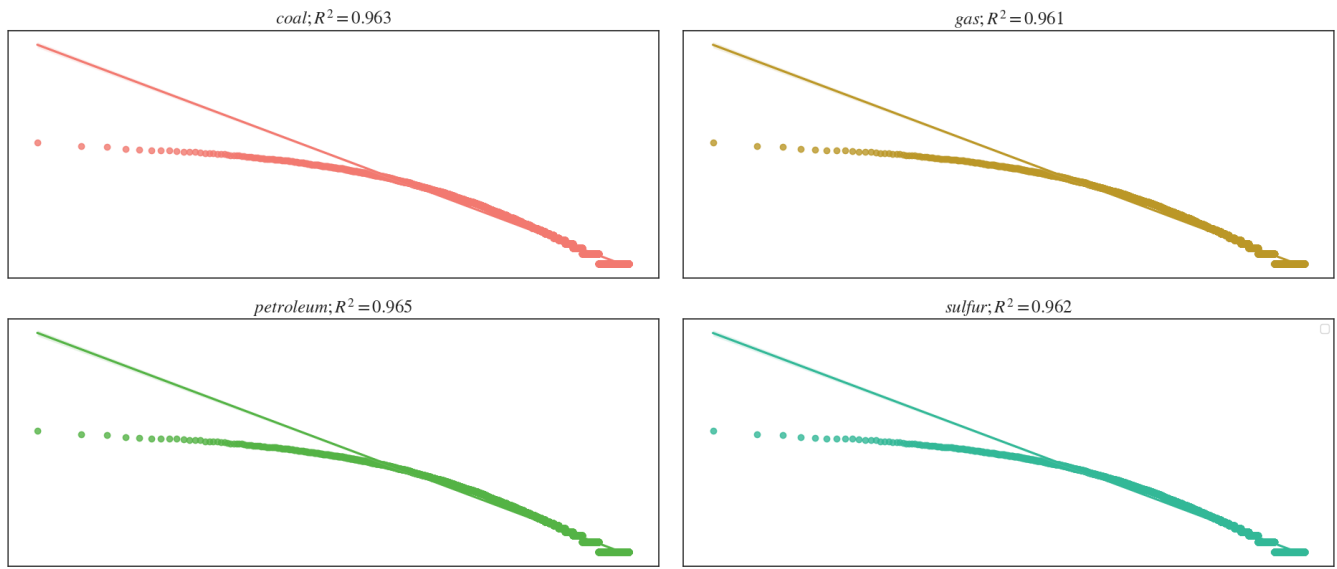

Fig.5: Frequency-rank curve ($\log_{10}$) of words in HAZOP reports.

## 4.2. Word classification

Firstly, the words are categorized as common or rare ones. Using the classical Zipf's law, the $f$-$r$ distribution parameters are calculated for the four report types, as detailed in Table 2. By analyzing the curvature, the respective $r_0$ values are determined as 239, 253, 153 and 199, as illustrated in Fig.6. Thus, common words with ranks ranging from 1 to $r_0$ and rare words with ranks ranging from $r_0+1$ to $r_t$ are identified.

Table 2: Parameters of Zipf's law.

| Type | Classical Zipf's law | | Extended Zipf's law | | |
|---|---|---|---|---|---|
| | α | C | α | C | β |
| petroleum | 0.965 | 51756 | 0.097 | 71.394 | − 81.000 |
| sulfur | 0.932 | 50040 | 0.087 | 64.852 | − 87.000 |
| Coal | 0.773 | 10829 | 0.116 | 62.278 | − 30.000 |
| Gas | 0.726 | 14194 | 0.094 | 68.330 | − 65.000 |

Table 2 also shows the parameters under the extended Zipf's law. The $r_0$ values for the four report types are calculated as 81, 87, 30 and 65 respectively. Similarly, common words and rare ones are identified.

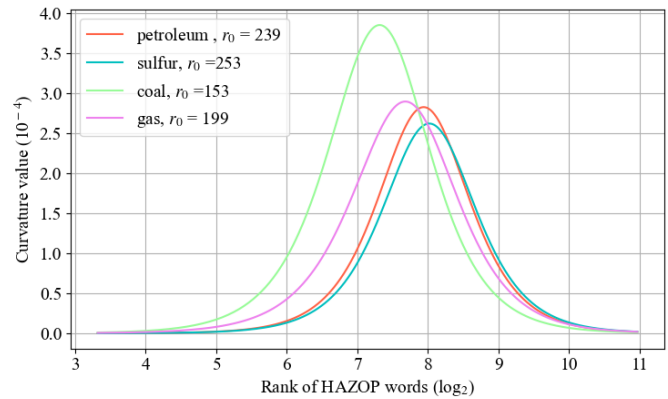

Fig.6: Curvature curve of classical Zipf distribution.

### 4.3. Sentence classification and generation

The sentence classification results are shown in Table 3, with illustrative examples included in Appendix 1.

Table 3: Sentence classification results.

| Type | Classical Zipf's law | | Extended Zipf's law | |
|---|---|---|---|---|
| | common | rare | common | rare |
| petroleum | 1603 | 3122 | 832 | 3893 |
| sulfur | 1585 | 3249 | 799 | 4035 |
| coal | 756 | 1481 | 455 | 1782 |
| gas | 1377 | 2051 | 776 | 2652 |

Sentence generation are performed separately by GPT2c and GPT2r models, as listed in Table 4.

Table 4: Quantity of inductive and expanded sentences.

| Type | Classical Zipf's law | | Extended Zipf's law | |
|---|---|---|---|---|
| | #inducive | #expanded | #inducive | #expanded |
| petroleum | 1603 | 6244 | 832 | 7786 |
| sulfur | 1585 | 6498 | 799 | 8070 |
| coal | 756 | 2962 | 455 | 3564 |
| gas | 1377 | 4102 | 776 | 5304 |

For instance, the sentence "When the flow is too small, prepare for standby linkage" is an inductive sentence generated based on the common sentences in Appendix 1. It's worth noting that an additional equipment entity called "Standby Linkage" is created, thereby enriching relevant entities.

Appendix 2 showcases the expanded sentences derived from rare sentence#2 in Appendix 1, where the relevant equipment and material entities are underlined. The first expanded sentence mainly focuses on the "pipeline" entity, potentially affected by the presence of "water pipe" in the original rare sentence. Thus, the contained entities relate to pipelines such as "oil supply pipeline", "gas pipeline" and "exhaust air pipeline". The second expanded sentence emphasizes the chemical compounds guided by "raw oil", such as "vinyl chloride", "hydrogen sulfide" and "sulfur dioxide", among others. Notably, some of these entities are new to the original rare sentence.

## 4.4. Rare entity labeling

To label material entities and equipment entities in the generated sentences, we employ POS analysis due to its suitability, and develop POS-dominated rule templates. For equipment entity labeling, the following approach is adopted for each generated sentence:

1. Use the Jieba tool to determine the POS of the sentence.

2. Locate the central word block $C_{wb}$ within the sentence, which can be an idiom (e.g., "the exhaust pipe"), or a noun group $ng_i$ (where $i$ can be equal to 1) with related nouns (such nouns and proper nouns) positioned next to each other, if any.

3. Examine the words above and below $ng_i$. If the last noun in $ng_i$ is followed by a quantifier, it should be included as part of $ng_i$. Otherwise, it should be discarded. If the first word of $ng_i$ is immediately followed by an adjective, it should be included in $ng_i$; otherwise it should be excluded.

4. Finally, label the $ng_i$ as an equipment entity.

Table 5: Examples of rare entity labeling.

| | |
|---|---|
| Generated sentences | **#1**: Liquid level of high alarm sulfur generation waste heat boiler is too low.<br>**#2**: Vinyl chloride, hydrogen sulfide and sulfur dioxide exceed the standard. |
| POS | **#1**: high (a) \| alarm (n) \| sulfur generation (n) \| waste heat (n) \| boiler (n) \| liquid level (n) \| too (ug) \| low (a)<br>**#2**: vinyl chloride (nz) \| ,(x) \| hydrogen sulfide (n) \| and (c) \| sulfur dioxide (nz) \| exceed the standard (v) |
| Labeling | **#1**: liquid (O) level (O) high (B-EQU) alarm (I-EQU) sulfur (I-EQU) generation (I-EQU) waste (I-EQU) heat (I-EQU) boiler (I-EQU) liquid level (I-EQU) too (O) low (O)<br>**#2**: vinyl (B-MAT) chloride （I-MAT), (O) hydrogen (O) sulfide (O) and (O) sulfur (B-MAT) dioxide (I-MAT) exceed (O) the (O) standard (O) |

Table 5 provides examples of rare entity labeling, where POS abbreviations are as follows: a (adjective), nz (other proper nouns), ug (tense particle), c (conjunction), and v (verb). In the first generated sentence, "high alarm sulfur generation waste heat boiler" is identified as an equipment entity. This sentence contains nested entities, such as "sulfur generation waste heat boiler". Using POS analysis, we determine that the $C_{wb}$, composed of noun group $ng_5$ (5 nouns), refers to the "alarm sulfur generation waste heat boiler liquid level". Next, we notice that the last noun "liquid level" in $C_{wb}$ is followed by a non-quantifier "over", indicating that "liquid level" should be discarded. Consequently, $C_{wb}$ becomes "alarm sulfur generation waste heat boiler". Additionally, since the first noun of $C_{wb}$ is immediately followed by the adjective "high", we retain it. Hence, the final equipment entity becomes "high alarm sulfur generation waste heat boiler". Finally, it is labeled "EQU" (i.e., equipment).

Regarding the labeling of material entities, we identify words in the generated sentences whose POS is classified as "other proper nouns". These words are then labeled as "MAT" (i.e., material), as shown in the second generated sentence in Table 5. Algorithm 2 outlines the pseudo code for the rare entity labeling process.

---

**Algorithm 2: Rare entity labeling**

**Input**: the generated sentence $g_s$
**Parameter**: i and q stand for the idiom and quantifier
**Output**: material entity $m_e$, equipment entity $e_e$

1:   $g_s \leftarrow$ POS
2:   **for** $p \leftarrow 1$ to len($g_s$) - 1 **do**
3:   **Case** I: // labeling of material entity
4:      **if** $g_s[p]$ is nz **then**
5:        $m_e \cup g_s[p]$
6:   **Case** II: // labeling of equipment entity
7:      **for** $q \leftarrow 1$ to len($g_s$) - 1 **do**
8:        **if** $g_s[p:q] \in \{n, i\}$ **then**
9:          $C_{wb} \leftarrow g_s[p:q]$
10:       **if** $g_s[p-1]$ is a **then**
11:         $C_{wb} \cup g_s[p-1]$
12:      **if** $g_s[q+1]$ is q **then**
13:        $C_{wb} \cup g_s[q+1]$
14:   $e_e \cup C_{wb}$
15:   **return** $m_e, e_e$

---

Note that the accuracy of the results can be verified by a series of auxiliary trigger words, such as "tank" and "pipeline" for equipment entities, "gas" and "grease" for material entities. Upon processing all the generated sentences, we successfully labeled 9,538 material entities and 9,629 equipment entities using classical Zipf's law, while 8,198 material entities and 7,733 equipment entities were labeled using extended Zipf's law. These labeled entities provide valuable supplements to the original HAZOP datasets. To ensure the reliability of the labeling process, three graduate students were invited to randomly select 1,000 entities for verification. The error rate determined through this verification process was 2.7%, which is considered acceptable.

## 5. EXPERIMENT

### 5.1. Dataset

Two datasets are created by preprocessing and manually annotating HAZOP reports (Wang et al., 2022a). The first dataset, called CPSYB, consists of reports related to **c**oal (specifically, the indirect coal liquefaction process) and **p**etroleum (specifically, the diesel hydrogenation process) obtained from **S**henhua Group, **Y**anshan Group and **B**runei group, called CPSYB dataset. The second dataset, named GSSL, comprises reports related to **g**as (specifically, the water electrolysis process) and **s**ulfur (specifically, the sulfur production process) obtained from **S**ichuan Petrochemical and **L**iaoyang Petrochemical. The entity annotation details can be found in Table 6.

Additionally, Fig.7 illustrates the distribution of entity frequency and its rank, which also demonstrates a linear relationship. This finding aligns with Zipf's law and highlights the presence of data imbalance within the datasets.

Table 6: Entity annotation strategy.

| Entity category | Initial | Subsequent |
|---|---|---|
| Equipment | B-EQU | I-EQU |
| Consequence | B-CON | I-CON |
| Material | B-MAT | I-MAT |
| State | B-STA | I-STA |
| Non-entity | O | O |

The CPSYB and GSSL datasets contain 78,315 and 72,189 lines of entities, respectively. Both datasets are randomly divided into sets for training, test and validation in the ratio of 8:1:1. Fig.7 illustrates the distribution of entity frequency and its rank, which also demonstrates a linear relationship. This finding aligns with Zipf's law and highlights the presence of data imbalance within the datasets.

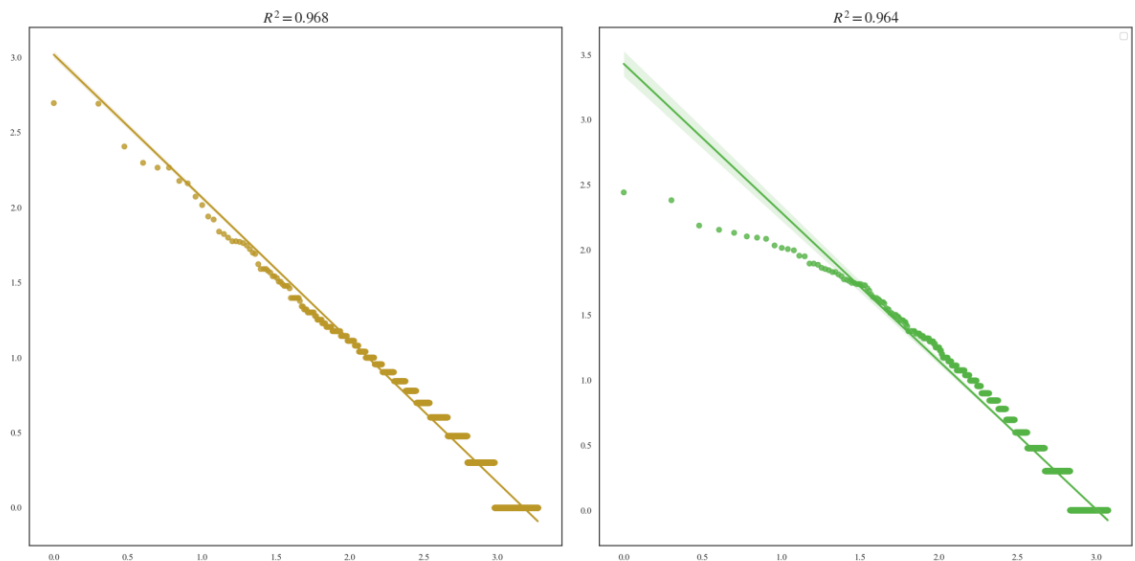

Fig.7: *f-r* distribution at the entity level (log$_{10}$ scale) on CPSYB (the left part) and GSSL (the right part).

**5.2. Trials**

To thoroughly evaluate our approach, we construct three widely-used entity extraction models (Liu et al., 2022): BiLSTM-CNN-CRF (Word2vec as the embedding layer), BERT-BiLSTM-CRF, and BERT-CNN-BiLSTM-CRF. These models have proven effectiveness in entity extraction across various domains, and are suitable for our purposes.

The Ziteg with the classical Zipf's law, denoted as Ziteg(c), is used to evaluate the effectiveness of the classical Zipf's law. The Ziteg with the extended Zipf's law, denoted as Ziteg(e), is used to evaluate the effectiveness of the extended Zipf's law. Moreover, we introduce the Ziteg with random sentence classification, denoted as Ziteg(r), to evaluate the gain of Zipf's law. All three Ziteg are applied to the aforementioned models, resulting in a total of 12 models: BiLSTM-CNN-CRF, BiLSTM-CNN-CRF-Ziteg(c), BiLSTM-CNN-CRF-Ziteg(e), BiLSTM-CNN-CRF-Ziteg(r), BERT-BiLSTM-CRF, BERT-BiLSTM-CRF-Ziteg(c), BERT-BiLSTM-CRF-Ziteg(e), BERT-BiLSTM-CRF-Ziteg(r), BERT-CNN-BiLSTM-CRF, BERT-CNN-BiLSTM-CRF-Ziteg(c), BERT-CNN-BiLSTM-CRF-Ziteg(e), and BERT-CNN-BiLSTM-CRF-Ziteg(r).

To provide further comparisons, we employ two popular re-sampling methods: random under-sampling (RUS) and random over-sampling (ROS). These methods are integrated into the three models, resulting in additional trials: BiLSTM-CNN-CRF-RUS, BERT-BiLSTM-CRF-RUS, BERT-CNN-BiLSTM-CRF-RUS, BiLSTM-CNN-CRF-ROS, BERT-BiLSTM-CRF-ROS, and BERT-CNN-BiLSTM-CRF-ROS.

Consistent parameters are employed for each trial, including Adam optimizer with 0.0003 learning rate, ReLU activation function, a batch size of 64, BERT-base model, and 50 epochs training. The reported results are the averages of five repeated experiments, and the evaluation metrics are F1 score, precision (P), and recall (R).

**5.3. Results**

**5.3.1. Effectiveness of Ziteg**

Table 7-8 show the model performance on the test set ("test") and validation set ("val") in terms of F1 for each entity category. The inclusion of Ziteg has resulted in improved performance for all models across all entity categories. This improvement is particularly noticeable in the GSSL dataset, especially with Ziteg(c) and Ziteg(e).

Table 7: Evaluation results (F1 %) for each category of entity on CPSYB.

| Model | equipment | | material | | consequence | | state | |
|---|---|---|---|---|---|---|---|---|
| | test | val | test | val | test | val | test | val |
| BiLSTM-CNN-CRF | 83.46 | 81.45 | 76.79 | 64.34 | 96.23 | 95.89 | 80.85 | 84.51 |
| BiLSTM-CNN-CRF-Ziteg(c) | 84.25 | 84.76 | 77.00 | 66.45 | 96.11 | 95.92 | 81.99 | 86.55 |
| BiLSTM-CNN-CRF-Ziteg(e) | 83.88 | 84.53 | 77.31 | 66.25 | 96.42 | 95.88 | 81.57 | 86.52 |
| BiLSTM-CNN-CRF-Ziteg(r) | 83.99 | 84.65 | 77.63 | 66.32 | 96.34 | 95.87 | 82.55 | 86.49 |
| BERT-BiLSTM-CRF | 85.40 | 85.36 | 77.06 | 67.54 | 96.30 | 96.33 | 83.67 | 88.37 |
| BERT-BiLSTM-CRF-Ziteg(c) | 87.66 | 87.92 | 80.32 | 70.10 | 98.74 | 98.08 | 85.84 | 90.33 |
| BERT-BiLSTM-CRF-Ziteg(e) | 87.07 | 87.66 | 79.65 | 68.05 | 97.97 | 97.24 | 85.12 | 90.01 |
| BERT-BiLSTM-CRF-Ziteg(r) | 86.38 | 85.61 | 78.19 | 67.55 | 96.88 | 96.47 | 83.00 | 88.38 |
| BERT-CNN-BiLSTM-CRF | 86.53 | 85.64 | 78.00 | 67.50 | 96.81 | 96.57 | 83.41 | 88.42 |
| BERT-CNN-BiLSTM-CRF-Ziteg(c) | 88.98 | 87.90 | 81.11 | 70.03 | 98.71 | 98.35 | 86.55 | 90.45 |
| BERT-CNN-BiLSTM-CRF-Ziteg(e) | 87.21 | 87.93 | 80.14 | 68.99 | 98.00 | 97.75 | 85.88 | 90.21 |
| BERT-CNN-BiLSTM-CRF-Ziteg(r) | 87.02 | 87.48 | 79.33 | 67.87 | 97.92 | 97.00 | 85.11 | 89.04 |

Table 8: Evaluation results (F1 %) for each category of entity on GSSL.

| Model | equipment | | material | | consequence | | state | |
|---|---|---|---|---|---|---|---|---|
| | test | val | test | val | test | val | test | val |
| BiLSTM-CNN-CRF | 46.15 | 36.41 | 59.26 | 28.96 | 30.04 | 23.53 | 76.40 | 40.00 |
| BiLSTM-CNN-CRF-Ziteg(c) | 46.77 | 44.13 | 65.12 | 37.25 | 32.50 | 27.26 | 79.52 | 44.44 |
| BiLSTM-CNN-CRF-Ziteg(e) | 45.95 | 42.47 | 66.50 | 37.09 | 37.50 | 25.44 | 76.70 | 42.93 |
| BiLSTM-CNN-CRF-Ziteg(r) | 47.33 | 43.19 | 68.82 | 36.88 | 41.00 | 26.05 | 78.05 | 41.37 |
| BERT-BiLSTM-CRF | 50.00 | 44.87 | 68.73 | 37.12 | 45.45 | 30.33 | 74.16 | 41.97 |
| BERT-BiLSTM-CRF-Ziteg(c) | 51.28 | 51.39 | 71.44 | 43.93 | 48.63 | 34.52 | 80.97 | 52.87 |
| BERT-BiLSTM-CRF-Ziteg(e) | 52.39 | 50.09 | 70.36 | 43.61 | 45.27 | 33.91 | 78.86 | 52.34 |
| BERT-BiLSTM-CRF-Ziteg(r) | 51.07 | 47.99 | 70.10 | 40.81 | 46.04 | 33.83 | 77.29 | 44.55 |
| BERT-CNN-BiLSTM-CRF | 40.00 | 44.55 | 64.94 | 37.88 | 42.11 | 28.97 | 80.95 | 41.76 |
| BERT-CNN-BiLSTM-CRF-Ziteg(c) | 51.33 | 53.74 | 72.74 | 44.25 | 48.05 | 33.91 | 81.25 | 53.32 |
| BERT-CNN-BiLSTM-CRF-Ziteg(e) | 55.29 | 49.66 | 71.05 | 44.87 | 57.14 | 34.42 | 82.35 | 52.11 |
| BERT-CNN-BiLSTM-CRF-Ziteg(r) | 53.33 | 48.33 | 71.35 | 39.52 | 49.64 | 33.99 | 81.19 | 46.01 |

Table 9-10 show the overall model performance on all entities. The inclusion of Ziteg has led to considerable gains for all three models. Notably, Ziteg(c) has elevated the F1 of BERT-BiLSTM-CRF and BERT-CNN-BiLSTM-CRF above 90%, and has demonstrated an over 10% recall increase on GSSL validation set. The increase in F1 is also noteworthy. Undoubtedly, Ziteg contributes to the enhancement of entity extraction performance.

Table 9: Evaluation results (%) on GSSL.

| Model | Precision | | Recall | | F1 | |
|---|---|---|---|---|---|---|
| | test | val | test | val | test | val |
| BiLSTM-CNN-CRF | 60.34 | 33.66 | 64.14 | 34.92 | 62.18 | 34.28 |
| BiLSTM-CNN-CRF- Ziteg (c) | 61.11 | 35.94 | 67.35 | 46.94 | 64.08 | 40.71 |
| BiLSTM-CNN-CRF- Ziteg (e) | 57.14 | 33.33 | 73.68 | 50.00 | 64.37 | 40.00 |
| BiLSTM-CNN-CRF- Ziteg (r) | 62.86 | 33.33 | 68.92 | 42.86 | 65.75 | 37.50 |
| BERT-BiLSTM-CRF | 65.29 | 39.26 | 71.01 | 41.46 | 68.03 | 40.33 |
| BERT-BiLSTM-CRF- Ziteg (c) | 68.79 | 47.42 | 75.72 | 55.90 | 72.09 | 51.31 |
| BERT-BiLSTM-CRF- Ziteg (e) | 67.23 | 46.25 | 72.75 | 55.06 | 69.88 | 50.27 |
| BERT-BiLSTM-CRF- Ziteg (r) | 65.77 | 38.89 | 72.50 | 46.65 | 68.97 | 42.42 |
| BERT-CNN-BiLSTM-CRF | 66.05 | 39.21 | 69.69 | 40.97 | 67.82 | 40.07 |
| BERT-CNN-BiLSTM-CRF- Ziteg (c) | 70.05 | 49.03 | 78.13 | 59.35 | 73.87 | 53.70 |
| BERT-CNN-BiLSTM-CRF- Ziteg (e) | 69.61 | 46.51 | 72.45 | 56.65 | 71.00 | 51.08 |
| BERT-CNN-BiLSTM-CRF- Ziteg (r) | 68.58 | 39.96 | 70.16 | 48.58 | 69.36 | 43.85 |

Table 10: Evaluation results (%) on CPSYB.

| Model | Precision | | Recall | | F1 | |
|---|---|---|---|---|---|---|
| | test | val | test | val | test | val |
| BiLSTM-CNN-CRF | 87.18 | 83.33 | 79.07 | 81.40 | 82.93 | 82.35 |
| BiLSTM-CNN-CRF-Ziteg(c) | 85.38 | 82.43 | 84.83 | 89.05 | 85.10 | 85.61 |
| BiLSTM-CNN-CRF-Ziteg(e) | 84.03 | 83.71 | 85.16 | 85.10 | 84.59 | 84.40 |
| BiLSTM-CNN-CRF-Ziteg(r) | 79.76 | 91.09 | 91.78 | 79.07 | 85.35 | 85.00 |
| BERT-BiLSTM-CRF | 85.44 | 86.03 | 87.83 | 87.71 | 86.62 | 86.86 |
| BERT-BiLSTM-CRF-Ziteg(c) | 90.87 | 88.77 | 90.01 | 92.44 | 90.44 | 90.57 |
| BERT-BiLSTM-CRF-Ziteg(e) | 88.76 | 87.83 | 89.34 | 89.30 | 89.05 | 88.56 |
| BERT-BiLSTM-CRF-Ziteg(r) | 88.50 | 87.64 | 87.03 | 86.43 | 87.76 | 87.03 |
| BERT-CNN-BiLSTM-CRF | 87.88 | 87.45 | 87.88 | 87.03 | 87.88 | 87.24 |
| BERT-CNN-BiLSTM-CRF-Ziteg(c) | 91.52 | 89.59 | 91.08 | 92.15 | 91.30 | 90.85 |
| BERT-CNN-BiLSTM-CRF-Ziteg(e) | 89.31 | 88.42 | 89.53 | 91.47 | 89.42 | 89.92 |
| BERT-CNN-BiLSTM-CRF-Ziteg(r) | 88.65 | 88.04 | 89.01 | 88.38 | 88.83 | 88.21 |

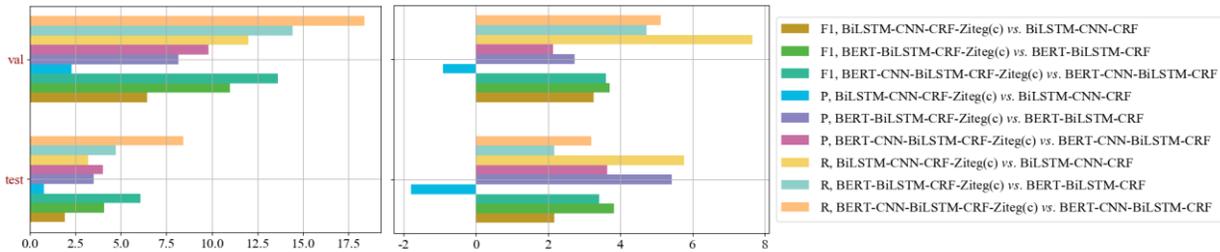

Fig.8: Performance gain through Ziteg(c).

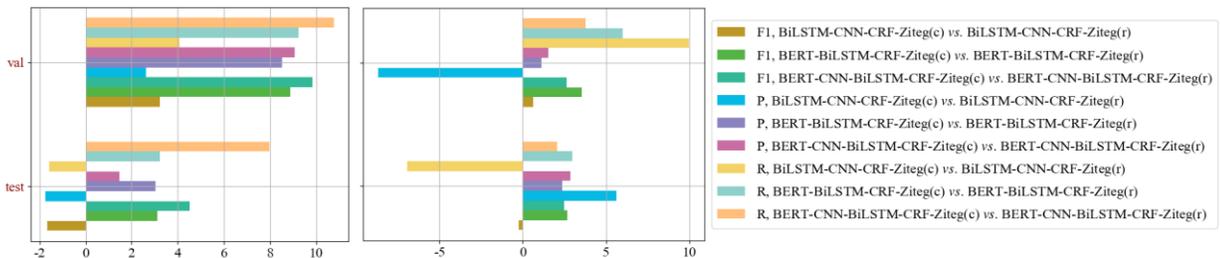

Fig.9: Performance gain of models with Ziteg(c) over models with Ziteg(r).

Furthermore, for Ziteg(c), Fig.8 illustrates its performance gains for GSSL and CPSTB datasets, displayed on the left and right respectively, following by similar figures. These figures visually demonstrate the performance difference between models. For example, BiLSTM-CNN-CRF-Ziteg(c) *vs.* BiLSTM-CNN-CRF indicates the extent to which the former model outperforms the latter. Similar comparisons can be made with other models using the respective figures.

Ziteg(c) exhibits positive improvements across all models, whether on GSSL test set or validation set, particularly in recall and F1 on the validation set, where the maximum performance gains surpass 18% and 13% respectively. On the CPSYB, Ziteg(c) also yields favorable results. Except for a slight decrease in precision for BiLSTM-CNN-CRF, other evaluations demonstrate gratifying results, with performance improvements of at least 2 percentage points. Fig.9 further confirms the superiority of Ziteg(c) over Ziteg(r). In conclusion, Ziteg(c) proves to be effective.

Fig.10 shows the impact of Ziteg(e). Obviously, Ziteg(e) also has a positive and intuitive stimulus on the models. With a few exceptions (i.e., the precision of BiLSTM-CNN-CRF on GSSL test and validation sets, precision of BiLSTM-CNN-CRF on CPSYB test set), all evaluations show improvement. Experimental results on GSSL dataset indicate that the recall performance has improved by over 11%, and the F1 has increased by more than 5%. On CPSYB dataset, the majority of evaluations have also improvements of more than 1%. Fig.11 further illustrates the advantages of Ziteg(e) over Ziteg(r). In conclusion, Ziteg(e) proves to be effective.

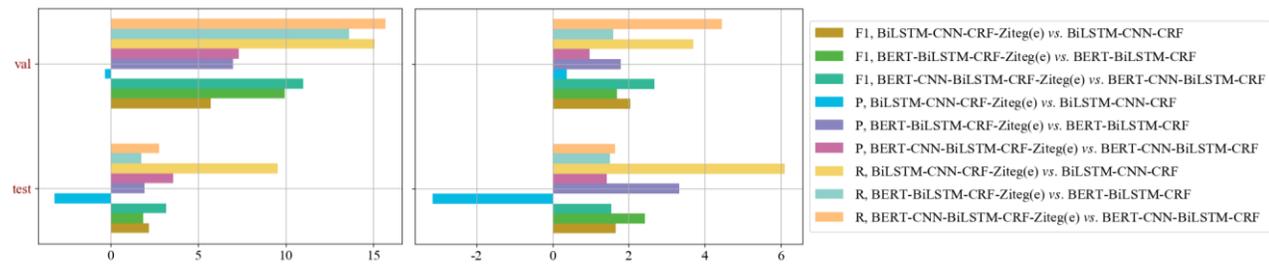

Fig.10: Performance gain through Ziteg(e).

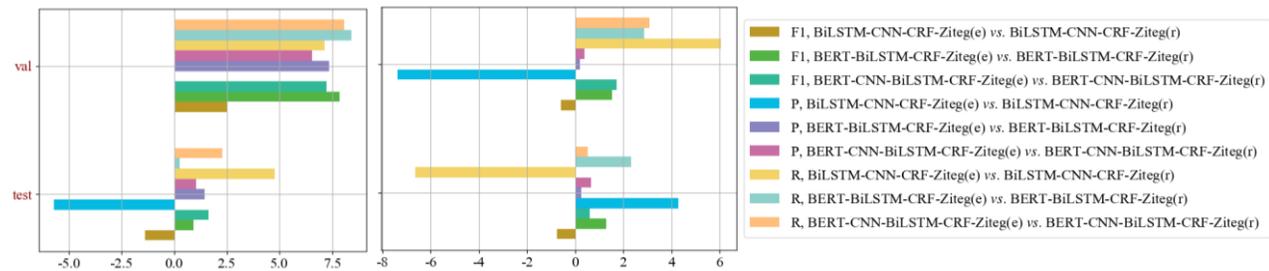

Fig.11: Performance gain of models with Ziteg(e) over models with Ziteg(r).

### 5.3.2. Sensitiveness of $r_0$

The parameter $r_0$ is crucial for the model performance. In both the classical Zipf distribution and the extended one, $r_0$ corresponds to the maximum curvature point and the turning point, respectively. While the theoretical and intuitive justifications for selecting $r_0$ are clear, it is beneficial to explore how the variation of $r_0$ affects the model's performance in order to fully leverage the benefits of Zipf's law.

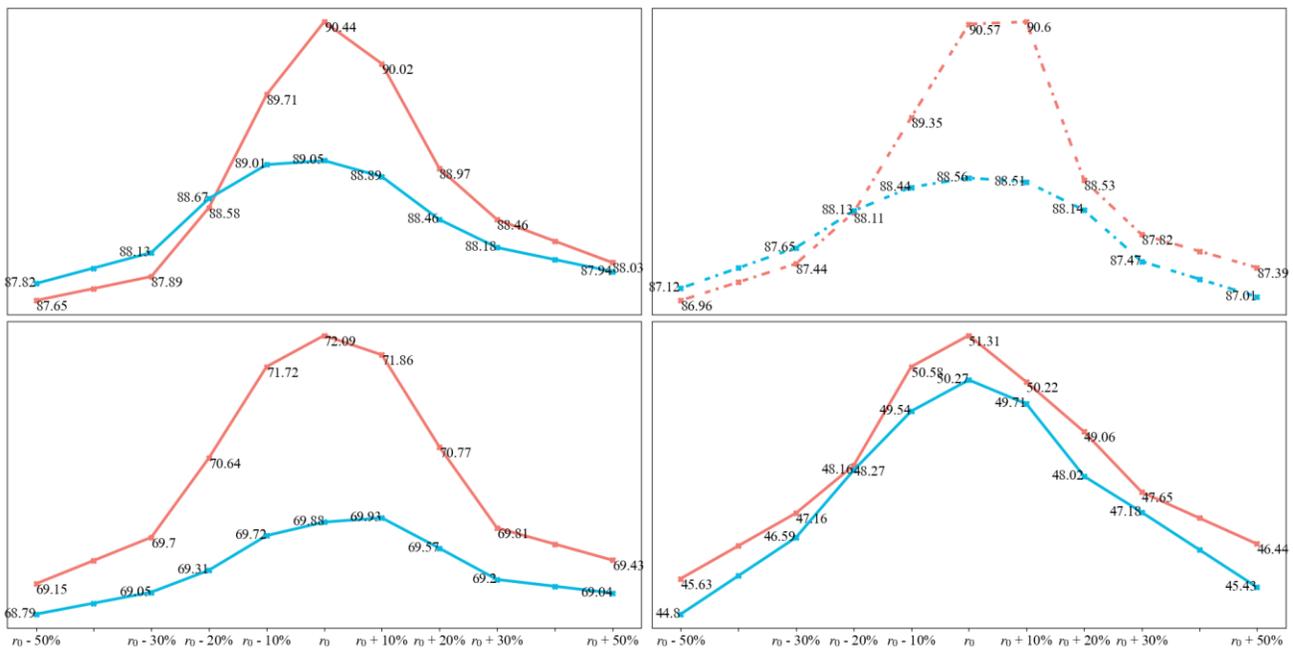

Fig.12: The F1 performance (%) trend of BERT-BiLSTM-CRF with 10%, 20%, 30%, and 50% parameters before and after $r_0$. Where, the red lines guide the classic Zipf's law, while the blue ones refer to extended Zipf's law.

We divide the values before and after $r_0$ into the deciles, and select parameters from the 10%, 20%, 30%, and 50% intervals before and after $r_0$. We then examine the impact of these parameters on the F1 performance of BERT-BiLSTM-CRF. Fig.12 displays the performance on CPSYB dataset (1st row) and GSSL dataset (2nd row), with the solid line for the test set and the dashed line for validation set.

From Fig.12, we can observe that the model's performance gradually declines as the parameter moves away from $r_0$, indicating that $r_0$ enables the model to achieve its best performance. The choices of the maximum curvature point and the inflection point can provide inspiration for researchers. However, there are two exceptions: a slightly higher F1 with $r_0 + 10\%$ using the classical Zipf's law on CPSYB validation set (indicated by the red dashed line in the top right), and a slightly higher F1 with $r_0 + 10\%$ using the extended Zipf's law on GSSL test set (indicated by the blue solid line in the bottom left). Despite these exceptions, $r_0$ remains the clear winner in the majority of other trials, enabling the model to achieve its best performance.

On the one hand, parameters after $r_0$ tend to dilute the proportion of rare words in rare sentences, resulting in fewer rare entities being generated. On the other hand, parameters before $r_0$ may cause rare sentences to miss critical rare words, resulting in the generated rare entities being insufficiently diverse. Thus, $r_0$ strikes a balance that allows the model to perform optimally.

### 5.3.3. Effectiveness comparison with RUS and ROS

Table 11 provides a result of Ziteg compared to RUS and ROS. Fig.13 showcases F1 on CPSYB test set and validation set in the first two images, followed by the performance for GSSL in the last two images. Overall, Ziteg demonstrates significant superiority, consistently outperforming the baselines to varying degrees.

Table 11: F1 performance (%) comparison between Ziteg and RUS, ROS.

| Model | CPSYB | | GSSL | |
|---|---|---|---|---|
| | test | val | test | val |
| BiLSTM-CNN-CRF-Ziteg(c) | 85.10 | 85.61 | 64.08 | 40.71 |
| BiLSTM-CNN-CRF-Ziteg(e) | 84.59 | 84.40 | 64.37 | 40.00 |
| BiLSTM-CNN-CRF-ROS | 83.57 | 82.99 | 62.85 | 37.44 |
| BiLSTM-CNN-CRF-RUS | 83.26 | 83.31 | 63.01 | 36.82 |
| BERT-BiLSTM-CRF-Ziteg(c) | 90.44 | 90.57 | 72.09 | 51.31 |
| BERT-BiLSTM-CRF-Ziteg(e) | 89.05 | 88.56 | 69.88 | 50.27 |
| BERT-BiLSTM-CRF-ROS | 86.93 | 87.22 | 68.26 | 43.13 |
| BERT-BiLSTM-CRF-RUS | 87.62 | 87.08 | 68.40 | 44.09 |
| BERT-CNN-BiLSTM-CRF-Ziteg(c) | 91.30 | 90.85 | 73.87 | 53.70 |
| BERT-CNN-BiLSTM-CRF-Ziteg(e) | 89.42 | 89.92 | 71.00 | 51.08 |
| BERT-CNN-BiLSTM-CRF-ROS | 87.98 | 87.66 | 68.50 | 43.32 |
| BERT-CNN-BiLSTM-CRF-RUS | 88.31 | 87.94 | 68.43 | 43.85 |

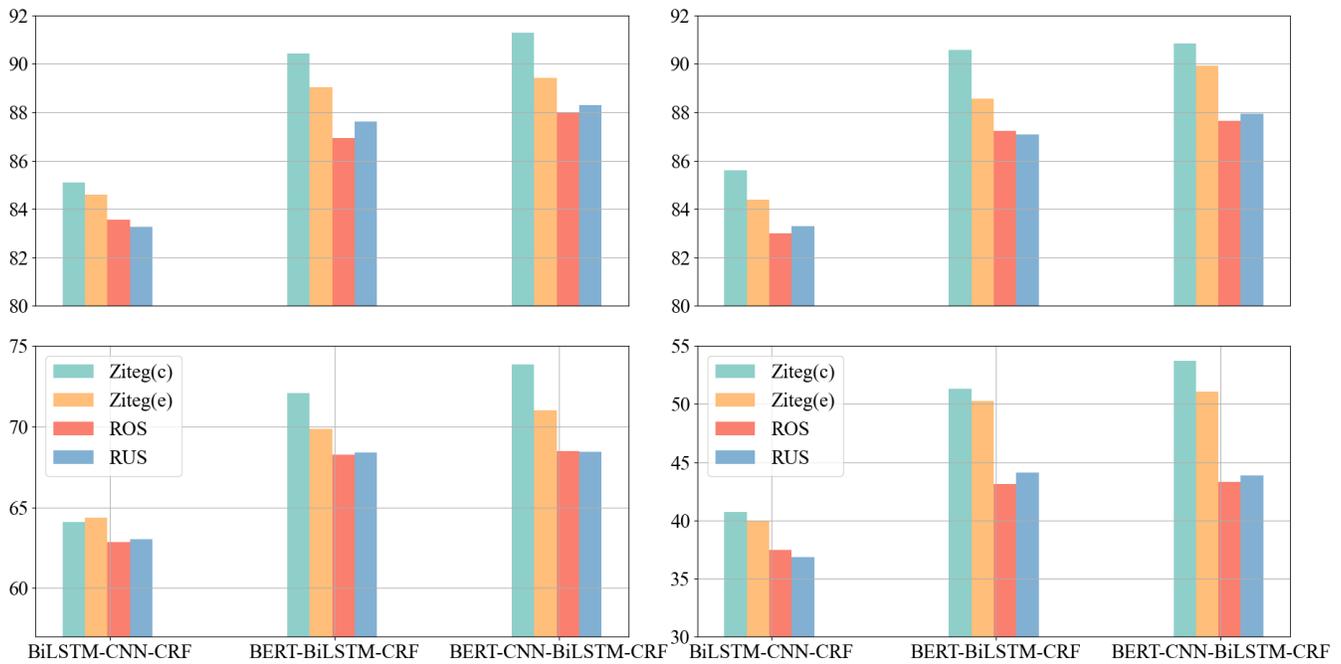

Fig.13: Comparison results with ROS and RUS.

On BiLSTM-CNN-CRF, Ziteg generally surpasses the baselines by at least 1%, outperforming both RUS and ROS in all four comparative experiments. This advantage is further evident on BERT-BiLSTM-CRF, particularly in GSSL. Ziteg(c) exhibits approximately 4% and 7% performance improvement over the baselines on the test and validation sets, respectively, while Ziteg(e) achieves remarkable 6% performance gain on the validation set. Similar remarkable improvements are observed on BERT-CNN-BiLSTM-CRF, where Ziteg notably outperforms the baselines, especially on GSSL validation set. This improvement can be attributed to Ziteg introducing additional stimulation.

### 5.3.4. Comparison of three Ziteg

Fig.9 and Fig.11 illustrate the significant superiority of Ziteg(c) and Ziteg(e) compared to Ziteg(r). We speculate that, in contrast to the randomness, Zipf's law embodies a determinacy that characterizes the dynamic nature of words, providing the models with additional a priori knowledge, thus enhancing their performance. Remarkably, Fig.14 depicts that on GSSL (left) and CPSYB (right), the model performance with Ziteg(c) outperforms that with Ziteg(e) in the majority of comparisons. This finding supports the simplicity principle inherent to Zipf's law, which emphasizes the attainment of greater benefits with minimal exertion. According to this principle, a single variable proves more effective than the inclusion of two variables. These insights are expected to inspire researchers to prioritize the classical Zipf's law in their future endeavors.

Consequently, the suggested priority emerges as follows: firstly Ziteg(c), followed by Ziteg(e), and finally Ziteg(r).

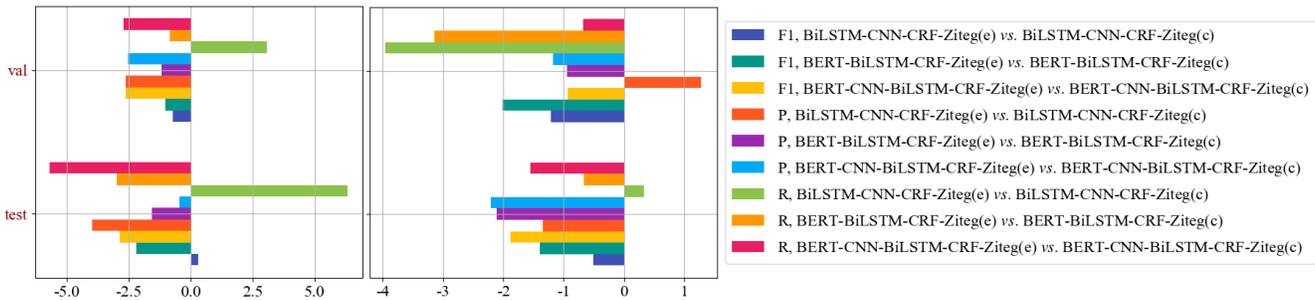

Fig.14: Performance gain of models with Ziteg(e) over those with Ziteg(c).

## 6. DISCUSSION

### 6.1. Informetrics' Active Role in Advancing AI Research

This article presents an intriguing study that underscores the active role of Informetrics in shaping the trajectory and advancement of AI, elucidating new avenues of inquiry and propelling the field forward. From the flow of knowledge perspective, this article serves as a conduit, to place the transmission of Informetrics knowledge into AI, accentuating the new idea of Zipf's law in guiding imbalance challenges, while amplifying its influence. To be specific, Zipf's law, which provides a valuable framework for understanding the distribution of word frequencies in natural language, is incorporated into text generation to enhance entity extraction tasks. Beyond this, Zipf's law holds significant value in AI research as it offers insights and principles that can be applied to various tasks like data analysis, modeling, and natural language processing.

The convergence of AI and Informetrics has gained increasing interest in recent years (Zhang et al., 2022). Informetrics provides profound insights on quantitative aspects of information, while AI enhances the endeavor through cutting-edge technologies like deep learning. The bibliometrics community has already demonstrated remarkable enthusiasm for integrating AI into broad information studies and transforming big data into valuable insights. Examples include bibliometrics-enhanced information retrieval (Mayr et al., 2014) and the extraction and evaluation of knowledge entities from scientific documents (Zhang et al., 2020).

Our study goes beyond the traditional grafting of AI onto Informetrics by showcasing the inherent value of Informetrics for AI, emphasizing its pivotal role in shaping the design and developmental potential of AI systems. By incorporating the foundational principles of Informetrics into AI, this article highlights the capacity to forge connections between disparate ideas, leading to discoveries that may elude a single disciplinary framework.

Effectively harnessing the power of AI and Informetrics to create cross-disciplinary solutions aligned with the big data boom requires further investigation from theoretical and practical perspectives. By synergistically integrating AI and Informetrics, researchers can navigate the big data landscape and proactively tackle the challenges and opportunities that arise in this era of massive data.

### 6.2. A closer analysis of Zipf's law

Zipf's law provides a macroscopic perspective to examine an entire corpus, but a closer analysis is necessary for a nuanced understanding. In this study, we aggregate words from HAZOP reports and divide them into four distinct frequency groups using quartiles. The resulting regression curves are presented in Fig.15.

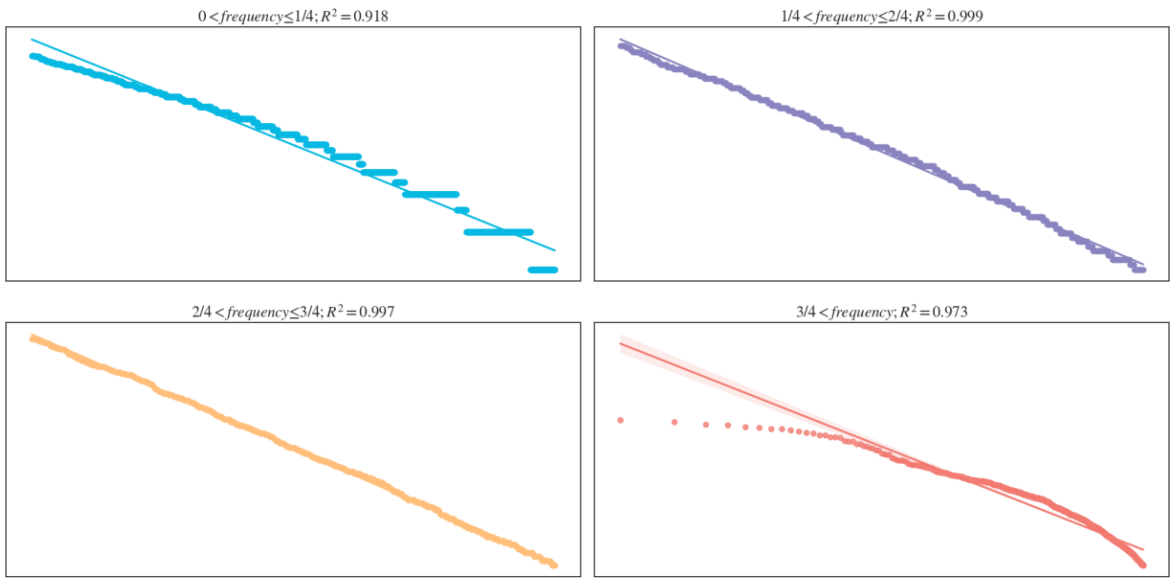

Fig.15: Word *f-r* curves below the quartile.

The regression fits remarkably well for the 25%-75% quartile, and well for the bottom quartile with 0.973 $R^2$. This indicates a strong correlation with Zipf's law, supporting its applicability to low-frequency and medium-frequency words in corpora. However, the regression for the top quartile (high-frequency words) falls slightly short, with 0.918 $R^2$. The dynamic nature of high-frequency word usage can be attributed to factors like syntax changes, contextual influences, and domain-specific terminologies.

The study suggests that lower-frequency word units exhibit less lexical activity, resulting in fewer variations in their usage. This observation indicates that low-frequency words tend to demonstrate greater stability in their semantic connotations and usage patterns, aligning more harmoniously with Zipf's law. Conversely, popular words with higher frequency are more prone to fluctuations due to their vigorous usage and exposure to contextual shifts. This divergence from strict adherence to Zipf's law becomes more evident in the case of high-frequency words.

Recognizing these patterns, it becomes crucial to prioritize the posterior portion of the Zipf's curve for analysis. This region offers indispensable insights and provides essential support for associated investigations. Additionally, scrutinizing the frequency distribution tail, which encompasses low-frequency words, grants supplementary perspectives into the salient features that permeate the corpus.

### 6.3. Fitting Zipf's law at the entity-level

The evolution of Zipf's law is rooted in the principle of least effort, suggesting that its manifestations extend beyond word frequencies to encompass the usage of entities as higher-level textual units. Fig.6 confirms the conformity of industry safety entities to Zipf's law. We also examine the distribution (*f-r*) between entity frequency and rank using an open dataset Resume NER (Zhang and Yang, 2018). The representation in the left of Fig.16 shows that the f-r does not exhibit a linear pattern ($R^2 = 0.750$). Further analysis reveals that entities with a frequency of 1 account for 83.80% (5159/6156), representing the least popular entities. Upon excluding these entities, the f-r distribution aligns more closely with Zipf's law, showing considerable improvement ($R^2 = 0.946$).

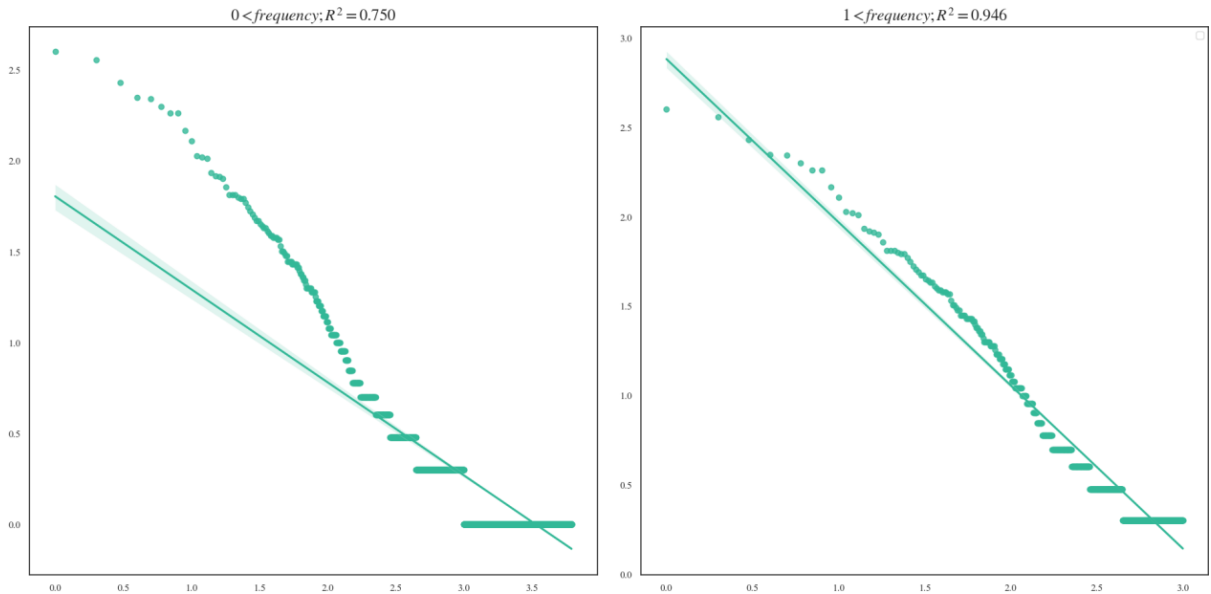

Fig.16: *f-r* fitting of Resume entities.

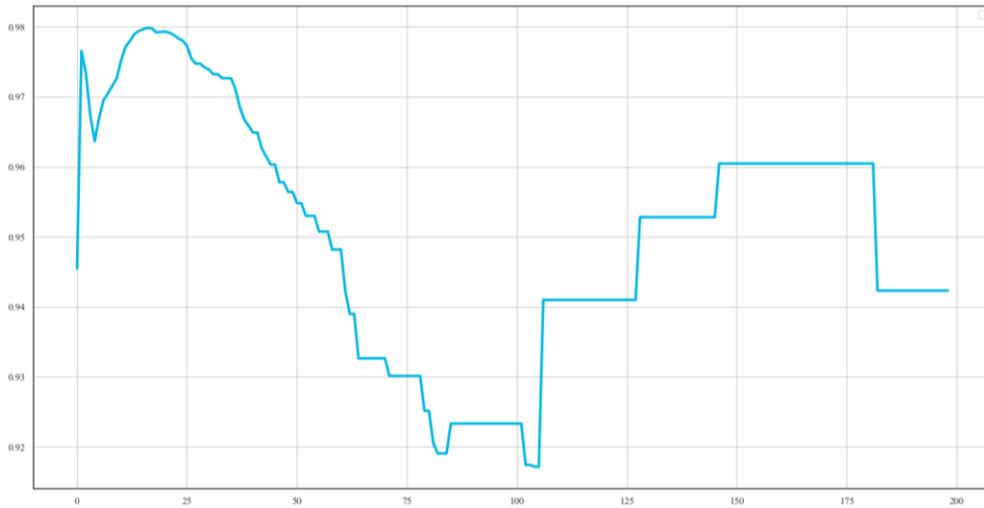

Fig.17: $R^2$ values of *f-r* fit of entities when progressively discarding the first *n* frequencies.

To delve deeper, we sort the entities based on their frequency in ascending order, gradually excluding the first *n* frequencies. By progressively fitting the *f-r* distribution of entities, we obtain the $R^2$ values, as shown in Fig.17. Notably, all subsequent Resume entities consistently exhibit a linear relationship, with the best fit achieved by excluding the first 25 frequencies ($R^2 > 0.97$), indicating strong adherence to Zipf's law. This finding suggests that Zipf's law can take on new forms, warranting further exploration through extensive statistical analysis. Our exploration of Zipf's law provides inspiration and insights for Informetrics studies, offering the potential to create greater value in practice.

## 7. CONCLUSION

This study introduces an innovative approach that leverages Zipf's law to address imbalance issue in entity extraction tasks. Under text generation techniques guided by Zipf's law, additional rare entities are generated and automatically labeled by classifying words and sentences, thereby mitigating the issue, and improving the entity extraction performance. The effectiveness of our approach is demonstrated through experiments on extracting entities from technical documents. Our findings highlight the superiority of the classical Zipf's law compared to the extended one. Additionally, we discuss the active role of informetrics in AI research and more details of Zipf's law.

This study acknowledges the potential limitations associated with model overfitting due to the introduction of auxiliary data and the inherent errors that may arise from automatically labeled samples, which could potentially weaken the model's performance. For instance, the performance of BiLSTM-CNN-CRF-Ziteg(c) is observed to be 0.12% lower than that of BiLSTM-CNN-CRF on the CPSYB dataset, as shown in Fig.10 and Fig.12. Addressing these concerns will be crucial for future research endeavors, aiming to refine and enhance the proposed approach.

## APPENDIX

Appendix 1: Examples of common sentences and rare sentences.

| Type | Classical Zipf's law | | Extended Zipf's law | |
|---|---|---|---|---|
| | common sentence#1 | rare sentence #1 | common sentence #2 | rare sentence #2 |
| petroleum | If the flow is too small, the fault will occur, the air will enter the furnace along the pipeline, the furnace temperature will decrease, and the flow will be interrupted in serious cases - execute shutdown, interlock, low alarm, and standby fan tail gas incinerator. | The feed oil pressurization causes the pump to stop due to failure (self-protection of electrical, mechanical and oil circuits is realized). In serious cases, the tank is full, and the feed oil enters the flare line, which is easy to freeze in winter. The liquid level of filtered feed oil buffer tank decreases, and the standby pump alarms when it is high. | High liquid level - circuit fault, small opening or closing - tank liquid level rises, pressure rises, and in severe cases, the gas phase breaks through the water seal and is discharged to the atmosphere. | Cracking and leakage of the heat tracing water supply pipe of the feed oil buffer tank may cause the feed oil to carry water into the reactor, resulting in catalyst breakage, and in serious cases, the unit will be shut down. |

Appendix 2: Examples of expanded sentences.

| Rare sentence | Expanded sentences |
|---|---|
| #1 | If the water pipe is seriously blocked, the leakage will seriously affect the life of the oil supply pipe. Section IV What should be paid attention to in gas leakage accidents? Why do gas leakage accidents occur frequently? Generally speaking, gas leakage and detonation mainly originate from the fire (fire explosion) at the pipeline interface, water leakage of gas pipeline, blockage of gas pipeline, and blockage of gas pipeline and water pipe. Section V Gas Leakage Prevention Measures After gas leakage occurs, in case of gas leakage and fire, it is necessary to timely find out, alarm and block whether the gas leakage facilities such as exhaust pipes and solenoid valves have been closed or stopped. Section VII Preventive Measures for Gas Leakage Alarm shall be given in time when gas leakage occurs. If gas leakage occurs, emergency repair and maintenance shall be carried out. If gas leakage occurs, isolation can be launched; The exhaust valve and transformer are subject to. |
| #2 | Minor corrosion such as vinyl chloride is the main cause of explosion. The main hazards of this corrosion are: 1) a large amount of corrosion, causing no oil out of the heater, serious explosion accidents, and even fire; 2) In serious cases, the gas equipment may be burnt, and in serious cases, the machine may be out of control! 3) Improper heating materials are added during processing, which cannot be guaranteed to be completely correct and cannot be discharged quickly. 4) The burning of coal may also damage the gas channel, causing fire or pollution of harmful and toxic gases such as oil products, oxidants, carbon dioxide, sulfur dioxide, etc. 5) Engine motor damage and engine scrap caused by corrosion; Hydrogen sulfide, sulfur dioxide, disulfide and other harmful substances seriously exceed the standard; Fire caused by excessive discharge; The above contents include the normal temperature limit of fuel storage temperature and heating temperature. |